\documentclass[letterpaper, 10 pt, journal, twoside]{IEEEtran}

\usepackage{graphics} % for pdf, bitmapped graphics files
\usepackage[normalem]{ulem}
\usepackage{xcolor}
\usepackage{float} % For [H]
\usepackage{stfloats} % For better float placement of wide figures
\usepackage{array}
\usepackage{epsfig} % for postscript graphics files
\usepackage{bm}
\usepackage{graphicx}
\usepackage{amsmath} % assumes amsmath package installed
\usepackage{amssymb}  % assumes amsmath package installed
\usepackage{cite}
\usepackage{booktabs}
\usepackage{siunitx}
\usepackage{algorithm}
\usepackage{algorithmicx}
\usepackage{algpseudocode}
\usepackage{multirow}
\usepackage{diagbox} 
\usepackage{hyperref}
\usepackage{relsize}
\usepackage{caption}
\ifCLASSINFOpdf
\else
\fi

\begin{document}
\title{Continual Learning for Traversability Prediction with Uncertainty-Aware Adaptation}

\author{Hojin Lee,  Yunho Lee, Daniel A Duecker, and Cheolhyeon Kwon%

\thanks{This work was supported in part by the National Research Foundation of Korea (NRF) grants (RS-2024-00342930 and RS-2020-NR049569), and in part by the InnoCORE program (N10250155(00)), both funded by the Korea government (Ministry of Science and ICT). \textit{(corresponding author: Cheolhyeon Kwon)}} %Use only for final RAL version
\thanks{Hojin Lee, Yunho Lee, and Cheolhyeon Kwon are with the Ulsan National Institute of Science and Technology, Ulsan, 44919, Republic of Korea {\tt\footnotesize (hojinlee@unist.ac.kr, dbs2911@unist.ac.kr, kwonc@unist.ac.kr)}.} 
\thanks{Daniel A Duecker is with Munich Institute of Robotics and Machine Intelligence (MIRMI), Technical University of Munich (TUM), Germany {\tt\footnotesize (daniel.duecker@tum.de)}.}
}

\maketitle

\begin{abstract}
Traversability prediction is a critical component of autonomous navigation in unstructured environments, where complex and uncertain robot-terrain interactions pose significant challenges such as traction loss and dynamic instability. Despite recent progress in learning-based traversability prediction, these methods often fail to adapt to novel terrains. Even when adaptation is achieved, retaining experience from previously trained environments remains a challenge, a problem known as catastrophic forgetting. To address this challenge, we propose a continual learning framework for traversability prediction that incrementally adapts to new terrains using a generative experience recall model. A key virtue of the proposed framework is two folds: i) retain prior experience without storing past data; and ii) incorporate the uncertainty of the generated samples from the recall model, enabling uncertainty-aware adaptation. Real-world experiments with a skid-steering robot validate the effectiveness of the proposed framework, demonstrating its ability to adapt across a series of diverse environments while mitigating catastrophic forgetting. 
\end{abstract}

% \begin{IEEEkeywords}
% IEEE, IEEEtran, journal, \LaTeX, paper, template.
% \end{IEEEkeywords}

\IEEEpeerreviewmaketitle

\section{Introduction}
\IEEEPARstart{A}{utonomous} navigation in unstructured environments poses significant challenges due to varying terrain traversability, making traversability prediction a central research focus \cite{sevastopoulos2022survey}. Traditionally, traversability prediction has been approached by crafting rules based on terrain geometry \cite{ahtiainen2017normal} and semantic context \cite{kim2006traversability}. However, these methods often fail to capture complex robot-terrain interactions. Recently, learning-based methods have offered promising solutions by leveraging driving data without relying on handcrafted traversability rules \cite{gasparino2022wayfast}.

While the learning-based traversability prediction methods perform well in conditions similar to the training environment, they often generalize poorly when deployed in novel terrains \cite{cai2025pietra}. To address this limitation, adaptation techniques have been proposed to update traversability models when encountering novel terrain. \cite{frey23fast}. However, such adaptation often leads to catastrophic forgetting, where newly acquired experience overwrites prior experience, resulting in degraded performance in previously trained environments.

To facilitate adaptation without catastrophic forgetting, strategies such as regularization-based continual learning and dynamic architectural methods have been explored \cite{wang2024comprehensive}. Among these methods, experience replay-based continual learning has been widely adopted in traversability learning, as it enables models to update by reusing past data without requiring complex optimization objectives or architectural changes \cite{ma2024imost,yoon2024adaptive}. These methods incrementally expand memory or selectively store data for experience replay. However, they are often hindered by high memory costs and/or the potential loss of valuable information during data selection.

\begin{figure}[tpb]
    \centering
    \includegraphics[width=0.45\textwidth]{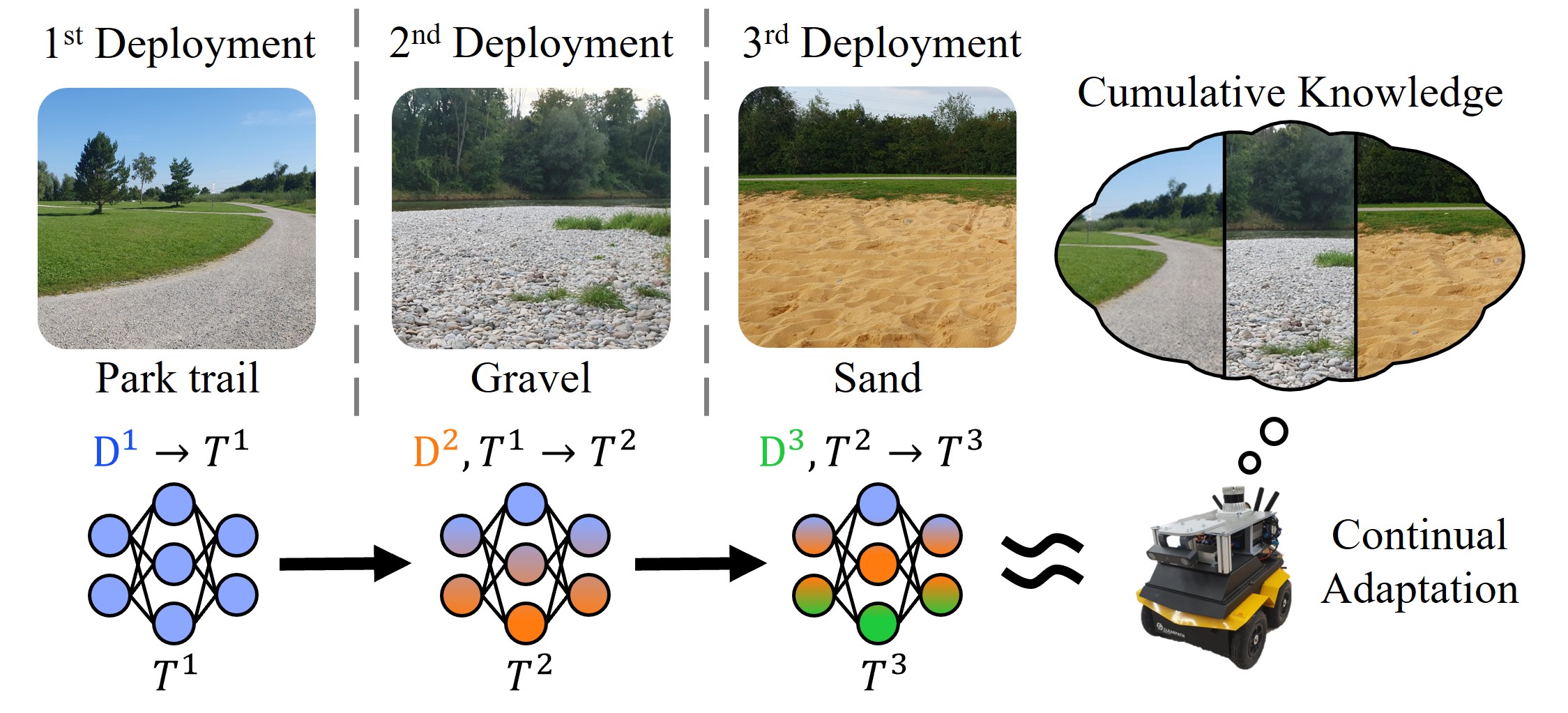}\caption{Continual adaptation across different types of terrains without access to past data. The robot incrementally updates its model while retaining experience for reliable traversability prediction.\vspace{-6mm}}\label{fig:main}
\end{figure}

To enable adaptation without losing prior experience and without relying on memory-demanding replay mechanisms, we propose a novel continual learning framework for traversability prediction that features uncertainty-aware adaptation. This framework integrates sensory environmental information and robot state information as input modalities, effectively learning to predict both traversability and its associated uncertainty. A key attribute of the proposed framework is a generative experience recall model, which retrieves prior experience to guide continual updates of both traversability prediction model and itself. In this way, prior experience can be retained without needing memory for storing past data. The generative recall model is further empowered by an uncertainty-aware adaptation mechanism, whereby the uncertainties of recalled samples are evaluated so that trustworthy samples are used to update the recall model. Building on this foundation, the trained traversability prediction model is integrated into a model predictive control (MPC) framework to navigate a robot across a series of diverse environments. 

These design choices enable a continual learning framework for traversability prediction that exploits a generative experience recall model for effective (i.e., retention of prior experience), efficient (i.e., no need to store past data), and reliable (i.e., uncertainty-aware) update. Real-world experiments with a skid-steering robot confirm the effectiveness of the proposed framework, showing its adaptability across diverse environments while retaining performance in previously trained environments. The main contributions are as follows: 
\begin{itemize}
        \item We propose an uncertainty-aware traversability prediction method that associates terrain features with the robot dynamics to capture complex robot-terrain interactions.
    \item We introduce a continual learning framework leveraging generative experience recall and an uncertainty-aware adaptation mechanism to retain prior experience while adapting to novel environments.    
    \item Real-world experiments using a skid-steering robot validate the proposed framework, demonstrating persistent navigation performance while learning uncertain traversability across diverse environments.
\end{itemize}

\vspace{-3mm}
\section{Related Work}
\begin{table*}[!htpb] 
\caption*{}
\caption{Comparison of Learning-based Traversability Prediction Methods with Adaptation Capability.} 
\label{table.related}
\centering
\renewcommand{\arraystretch}{1.1} % Increase row spacing
\begin{tabular}{l|cccc} 
\hline
 & 
\begin{tabular}[c]{@{}c@{}}Input modality\end{tabular} & 
\begin{tabular}[c]{@{}c@{}}Adaptation technique\end{tabular} & 
\begin{tabular}[c]{@{}c@{}}Prior-experience memory\end{tabular} & 
\begin{tabular}[c]{@{}c@{}}Uncertainty modeling\end{tabular} \\ 
\hline
\cite{frey23fast} & Vision & Adaptation with recent experience & N/A & Epistemic \\ 
\cite{yoon2024adaptive} & Vision & Experience replay & Fixed-sized memory & Aleatoric \\ 
\cite{ma2024imost} & Vision & Experience replay & Incremental dynamic memory & - \\ 
Proposed & Vision, geometry, dynamics & Generative experience replay & N/A & Aleatoric, epistemic \\ 
\hline
\end{tabular}\vspace{-5mm}
\end{table*}

\subsection{Learning-based Traversability Prediction}
Learning-based methods have emerged as a powerful alternative to traditional rule-based methods in traversability prediction. They rely on real or simulated data to model complex robot-terrain interactions with minimal supervision \cite{ahtiainen2017normal}. A critical design choice in prediction models is the input data modality, which directly impacts the model's ability to comprehend underlying dynamics. Geometry features are commonly used \cite{ahtiainen2017normal}, but they are insufficient to capture semantic terrain characteristics. To address this, the Learning Applied to Ground Vehicles (LAGR) program exerted pioneering effort on employing visual features for traversability prediction \cite{kim2006traversability}.

Learning terrain geometry and visual features may not be sufficient, since traversability is substantially influenced by the robot’s state and dynamics. Some methods attempt to exploit proprioceptive sensor data to capture this influence, such as measuring traversability in terms of vibration level \cite{wellhausen2019should}. However, they still fall short in fully representing traversability, making it difficult to assess the admissible states for navigating in a given terrain. To comprehend the interdependencies between the robot’s state and terrain features in traversability prediction, \cite{pokhrel2024cahsor} encompasses the robot's speed, inertial sensor data, and terrain visual information to establish a forward kinematics model. This model can predict the robot's dynamic response on the terrain, interpreting its traversability. A major limitation of this approach, however, is the lack of uncertainty modeling, which degrades the robustness and reliability of traversability predictions. Furthermore, the aforementioned learning-based methods often fail to generalize when deployed in environments that differ from the training conditions. In response to this shortcoming, we propose a continual learning-based framework that combines terrain features and robot state information for traversability prediction, while also taking into account uncertainty.

\subsection{Adaptation for Traversability Prediction}
Learning-based traversability prediction faces challenges in novel environments that differ from the trained environment. Adaptation methods address these challenges by updating the model with newly collected data, focusing on rapid adaptation to new environments \cite{frey23fast}. However, they overlook retaining experience from previously trained environments during updates. This leads to degraded navigation performance when the robot is deployed in previous environments.

Experience replay mechanisms have been proposed to address this issue by storing past data in a memory buffer and replaying it alongside newly acquired data \cite{wang2024comprehensive}. \cite{ma2024imost} employs an incremental replay buffer that clusters visual features, but its memory footprint grows with environmental diversity, limiting scalability in long-term deployment. In contrast, \cite{yoon2024adaptive} uses a fixed-size replay buffer, carrying the risk of discarding valuable information when adding new data. These methods present a fundamental trade-off between scalability and representational capacity, making it difficult to retain prior experience across diverse environments. Moreover, neither \cite{ma2024imost} nor \cite{yoon2024adaptive} models epistemic uncertainty, which is essential for assessing prediction confidence in unfamiliar environments. They also primarily rely on terrain appearance while neglecting the influence of robot dynamics on traversability, limited in capturing the complexity of robot–terrain interactions.

To address the shortcomings in experience replay mechanisms, generative model-based continual learning offers an alternative by synthesizing past data from learned representations, thereby eliminating the need to store past data and supporting scalable adaptation to diverse environments \cite{wang2024comprehensive}. While offering notable advantages, such methods often fail to account for the uncertainty associated with generated samples, which undermines the prediction model’s updates and potentially compromises performance. To this end, we propose a framework that accounts for the uncertainty of the generated samples and facilitates reliable adaptation to new data and retention of prior experience.

To provide a comparative overview, Table \ref{table.related} summarizes key aspects of the representative traversability prediction methods with adaptation, including the proposed framework.

\begin{figure*}[t]
\centering
\includegraphics[width=0.9\linewidth]{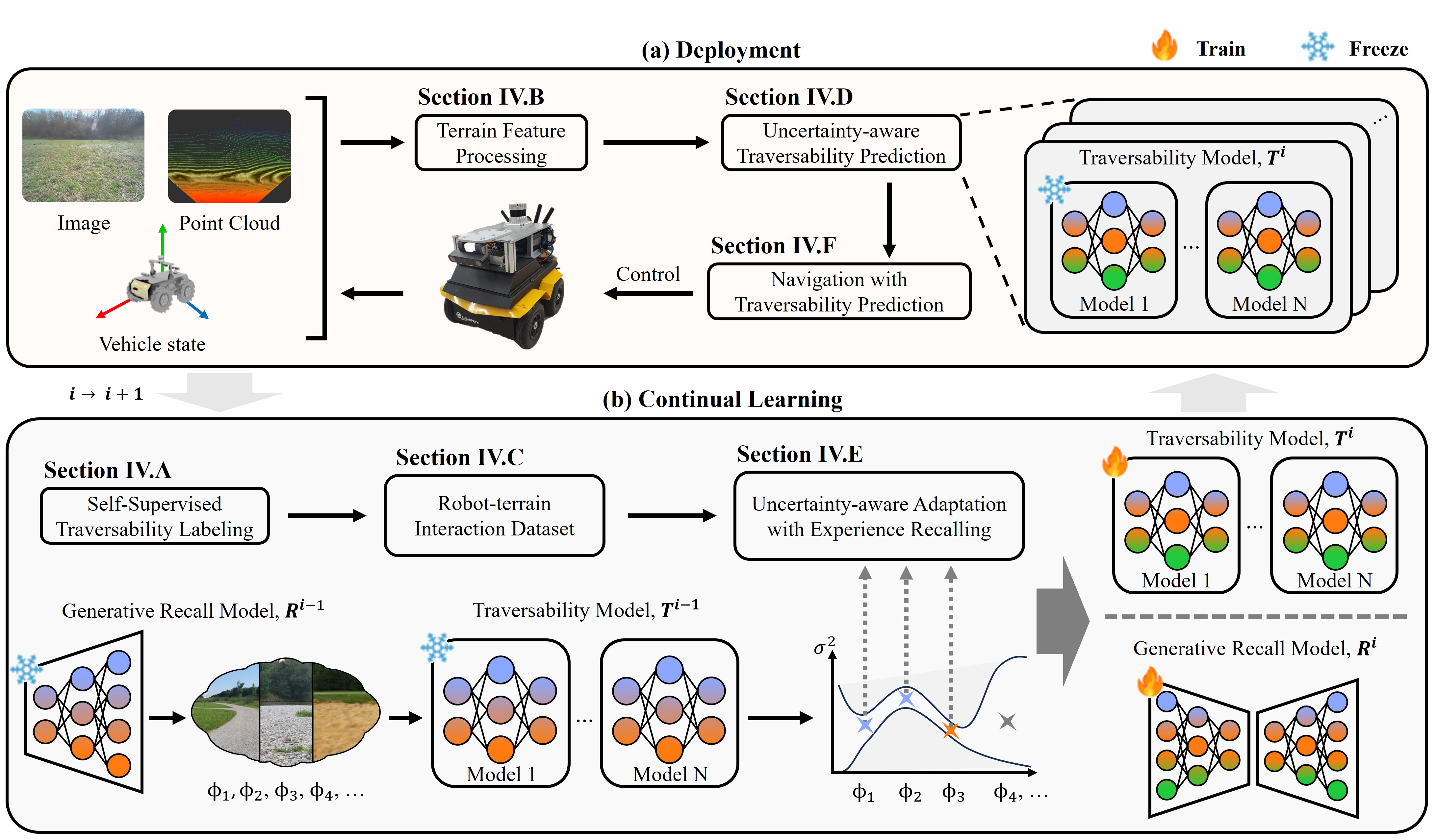}
\caption{Overview of the proposed framework. (a) Deployment phase: uncertainty-aware traversability prediction and navigation. (b) Continual learning phase: uncertainty-aware model adaptation to new environments via generative experience recall.} \vspace{-5mm}
\label{fig:main_alg_row}
\end{figure*}

\section{Preliminaries}
\subsection{Robot Dynamics with Traction Parameters}\label{ssc:dynamics}
Let us consider a discrete-time dynamics for a ground robot described by:
\begin{equation*}
    x_{k+1} = f(x_k, u_k, \xi_k),
\end{equation*}
where \(x_k\) represents the robot's state, \(u_k\) is the control input, and \(\xi_k\) denotes the traction parameter at time step \(k\), accounting for variability in robot-terrain interactions. $f$ is modeled by the following unicycle-based kinematic as it effectively captures the dynamics of a wide range of ground robots, including skid-steering and legged platforms:
\begin{equation}\label{eq:dyn}
    \begin{split}
        \begin{bmatrix}
            p_{k+1}^x \\
            p_{k+1}^y \\
            \theta_{k+1}
        \end{bmatrix}
        = 
        \begin{bmatrix}
            p_k^x \\
            p_k^y \\
            \theta_k
        \end{bmatrix}
        + \Delta t \cdot
        \begin{bmatrix}
            \xi_k^x \cdot v_k^{cmd} \cdot \cos(\theta_k) \\
            \xi_k^x \cdot v_k^{cmd} \cdot \sin(\theta_k) \\
            \xi_k^z \cdot \omega_k^{cmd}
        \end{bmatrix},
    \end{split}
\end{equation}
where the state vector \(x_k = [p_k^x, p_k^y, \theta_k]^\top\) represents robot's \(X-Y\) position in global frame and its yaw angle. The control input \(u_k = [v_k^{cmd}, \omega_k^{cmd}]^\top\) consists of the linear and angular velocity commands, and \(\Delta t > 0\) denotes the sampling time. The traction parameters \(\xi_k = [\xi_k^x, \xi_k^z]^\top \in \mathbb{R}^2 \) account for the linear and angular discrepancies between the commanded and actual velocities due to uncertain robot-terrain interactions. Without loss of generality, we consider these discrepancies as a measure of traversability \cite{cai2024evora}.

\subsection{Planning with Uncertain Traversability Prediction}\label{ssc:navigation}
In unstructured environments with varying traversability, the navigation problem involves determining the optimal control input sequence $u_{k:k+T-1}^*$ that minimizes the robot’s navigation cost over a receding horizon of length $T$ \cite{cai2024evora}. Given the state $x_{k}$, and a navigation cost function $\mathcal{C}$, the receding horizon stochastic optimal control problem can be formulated as:
\begin{equation}\label{eq:obj}
\begin{split}
    \min\limits_{u_{0:T-1}} \ & \mathbb{E}[ \mathcal{C}(x_{0:T},u_{0:T-1})]\\
     \mathrm{s.t.} & \ \  \quad x_{0}=x_{k}, \\
    &\ x_{t+1}=f(x_{t},u_{t},\hat{\xi}_{t}),\    \forall t \in \{0,\cdots, T-1\}
\end{split}
\end{equation}
where $\hat{\xi}_{t} \sim \mathcal{P}(\xi|x_{t})$ represents the distribution of traction at time step $t$, which can be predicted by the traversability prediction model. To solve \eqref{eq:obj}, we utilize a sampling-based MPC method, leveraging its parallelizability on GPUs for real-time computation \cite{lee2023learning}.

\subsection{Continual Learning Problem Formulation}\label{ssc:problem_formulation}
To reliably codify $\hat{\xi}$ across different environments, we tackle the problem of domain-incremental continual learning for traversability prediction \cite{wang2024comprehensive}. We define the traversability prediction model \( \mathcal{T}: \mathbb{R}^q \rightarrow \mathbb{R}^2 \times \mathbb{R}_{>0}^2 \) as a mapping from input features, comprising the robot’s state and terrain features, to the mean and variance of the Gaussian distributions over traction parameters $\hat{\xi}\sim\mathcal{N}(\mu(\cdot), \sigma^2(\cdot))$, where $(\mu(\cdot), \sigma^2(\cdot))=\mathcal{T}(\cdot)$. The goal is to train a sequence of models $\mathcal{T}^1, \cdots,\mathcal{T}^{i-1}$, such that $\mathcal{T}^i$ adapts to new environments encountered during the $i^\text{th}$ deployment, while retaining experience from previous $(1^{st}, \cdots, {i-1}^{th})$ deployments. In this setting, the input distribution (environmental terrain characteristics) may shift over deployments, whereas the underlying task of learning (traversability prediction model) remains consistent.

At each deployment phase $i$, the robot collects data $\mathcal{D}^i$ in the corresponding environment, forming a sequential dataset $\mathcal{D}^{1:i} = \{ \mathcal{D}^1, \dots, \mathcal{D}^i \}$. Ideally, the model $\mathcal{T}^i$ would be trained on the full dataset $\mathcal{D}^{1:i}$, but storing and accessing all the past data is impractical on resource-constrained platforms. Instead, our continual learning framework aims to update $\mathcal{T}^{i-1}$ using only $\mathcal{D}^i$, with the goal of producing $\mathcal{T}^i$ that preserves performance without direct access to $\mathcal{D}^{1:i-1}$.

\section{Algorithm Development}
This section presents the proposed continual learning framework for traversability prediction, empowered by uncertainty-aware adaptation, as depicted in Fig. \ref{fig:main_alg_row}.

\subsection{Self-Supervised Traversability Labeling}\label{sc:mainlabeling}
To train the traversability prediction model, we collect samples of traction parameters $\xi$ while navigating over different terrain environments, including both traversable and non-traversable regions. Inspired by the nonlinear moving-horizon estimator in \cite{gasparino2022wayfast}, we utilize odometry to estimate the robot’s state $x$ and use this information to estimate the traction parameters $\xi$. Specifically, we employ an Extended Kalman Filter (EKF) where the traction parameters are modeled as a random walk process: $\xi_{k+1} = \xi_k + \nu$. Here, $\nu$ is zero-mean Gaussian process noise that accounts for gradual variations over time \cite{bloesch2013state}. Using \eqref{eq:dyn} along with state $x_{k+1}$, the EKF iteratively refines the traction parameter estimates $\xi_k$, establishing datasets that are used to train the traversability prediction model.

\subsection{Terrain Feature Processing}\label{sc:mainfeatprocess}
We use RGB images and 3D point clouds to build a multi-modal terrain representation based on \cite{erni23mem}. The point clouds are processed into an elevation grid map, capturing the geometric properties of the terrain. Simultaneously, visual features are extracted from images using a pre-trained DINOv2 backbone \cite{oquab2023dinov2}, producing pixel-wise feature embeddings. However, projecting these high-dimensional visual features (e.g., 384 dimensions for the DINOv2-small model) into a grid map incurs substantial computational and memory overhead, making it impractical for real-time use on mobile robots.

\vspace{-3mm}
\begin{figure}[H]
\centering
\includegraphics[width=0.45\textwidth]{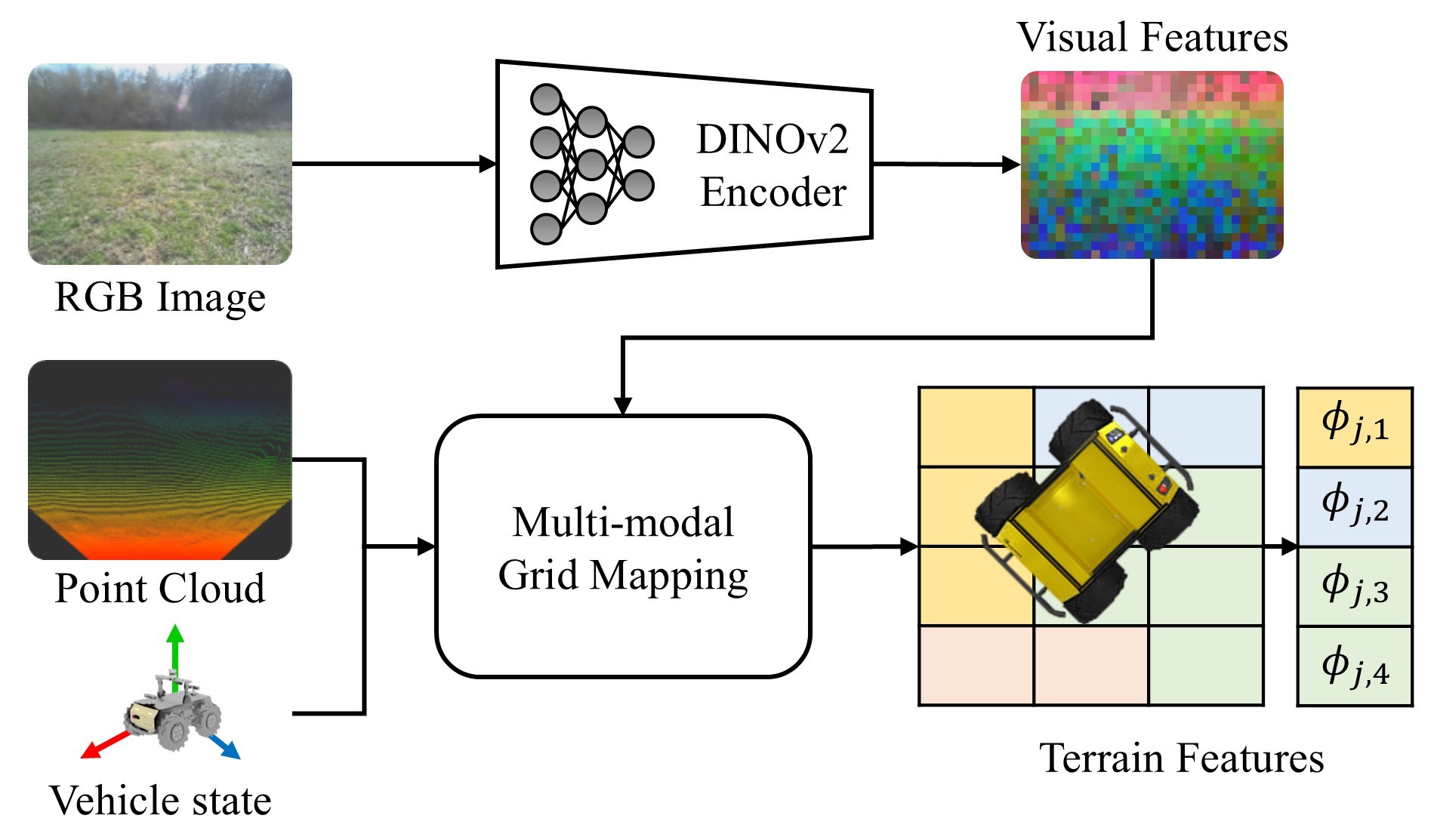}
\caption{Grid mapping of the terrain feature $\phi$, based on geometric and visual information of the corresponding terrain patches beneath the robot’s four wheels.\vspace{-3mm}}\label{fig:terrainfeature}
\end{figure}

To address this, we employ an autoencoder network to compress the high-dimensional visual feature embeddings from DINOv2 into a low-dimensional latent feature. The network is trained in an unsupervised manner on the original dense DINOv2 features from a large-scale off-road dataset by optimizing reconstruction loss. Once trained, the encoder part is deployed for real-time processing, receiving dense features from the DINOv2 backbone and generating compact, low-dimensional visual features to be integrated into the multi-modal grid map. As a result, terrain features $\phi \in \mathbb{R}^{p}$, 
implicating both geometric and visual information, are assigned to individual grids in the map, as illustrated in Figure \ref{fig:terrainfeature}. These terrain features are stored in a local grid map centered on the robot’s pose, capturing terrain characteristics over a fixed window for traversability model training and prediction.

\subsection{Robot-terrain Interaction Dataset}\label{sc:dataset}
Based on the labeled traction parameter data, driving data, and terrain feature data, we construct a robot-terrain interaction dataset:
\begin{equation*}
    \mathcal{D} = \left\{ (\phi_j, v_j, \xi_j) \mid j = 1, \dots, \mathcal{B} \right\},
\end{equation*}
where each sample consists of the terrain feature vector $\phi_j$, the robot’s velocity state $v_j$, and the labeled traction parameter $\xi_j$ at time step $j$. \( \mathcal{B} \) denotes the total number of samples in the dataset. The terrain feature vector $\phi_j = \left[ \phi_{j,1}^\top, \phi_{j,2}^\top, \phi_{j,3}^\top, \phi_{j,4}^\top \right]^\top$ consists of multiple terrain features extracted at different contact points. For instance, in a four-wheeled robot, each \( \phi_{j,i} \) corresponds to the terrain features at the $i^{th}$ wheel position $(p^{x}_{j,i}, p^{y}_{j,i})$ at time step $j$. The robot’s velocity state $v_j = \left[ v^x_j, w^z_j \right]^\top$ includes the linear and angular velocities estimated by odometry. To construct the dataset, we retrieve the stored local terrain grid maps for each recorded robot pose. For each map, we project the trajectory history within a local window and extract terrain features at the wheel contact points along the trajectory. These features are then paired with the corresponding robot velocity and traction estimates to form the dataset. Samples with missing terrain features at any wheel contact point due to being outside the sensor field of view are discarded.

\subsection{Uncertainty-aware Traversability Prediction}\label{sc:mainest}
Given the dataset $\mathcal{D}$, the model $\mathcal{T}$ is trained to predict the traction parameter as $\hat{\xi}_k = \mathcal{T}(\phi_k, v_k)$, where $\phi_k$ represents the terrain features, and $v_k$ denotes the robot velocity state at time step $k$. To account for uncertainty, we model $\mathcal{T}$ as a probabilistic ensemble model capable of capturing both aleatoric and epistemic uncertainties \cite{lee2023kino}. Each ensemble member receives the concatenated terrain feature $\phi_k$ and velocity state $v_k$, as input and is trained independently. All ensemble members share the same architecture, comprising a multi-layer perceptron with fully connected layers of sizes $[16, 32, 32, 16]$, each followed by a LeakyReLU non-linear activation function. The $m^{th}$ ensemble model $\mathcal{T}_m$ outputs the traction parameter as a Gaussian distribution:  
\begin{equation*}
    \hat{\xi}_{k,m} \sim N(\mu_{k,m}, \sigma_{k,m}^{2}), \ \ \text{where} \ [\mu_{k,m}, \sigma^2_{k,m}] = \mathcal{T}_m(\phi_k, v_k).
\end{equation*}
Based on predictions from $M$ ensemble models, we compute the final prediction as:
\begin{equation}\label{eq:ensemble_output}
    \hat{\xi}_{k} \sim N(\mu_{k}, \sigma_{k}^{2}), \ \  \text{where} \ [\mu_{k}, \sigma^2_{k}] = \mathcal{T}(\phi_k, v_k),
\end{equation}
with the mean and variance computed as:
\begin{equation*}
    \mu_{k} = \frac{1}{M} \sum_{m=1}^{M} \mu_{k,m},  
    \quad \sigma^2_{\text{k}} = \sigma^2_{\text{k,ale}} + \sigma^2_{\text{k,epi}},
\end{equation*}
where the variances from aleatoric $\sigma^2_{\text{k,ale}}$ and epistemic $\sigma^2_{\text{k,epi}}$ uncertainties are respectively given by:
\begin{equation*}
    \sigma^2_{\text{k,ale}} = \frac{1}{M} \sum_{m=1}^{M} \sigma^2_{k,m}, 
    \quad \sigma^2_{\text{k,epi}} = \frac{1}{M} \sum_{m=1}^{M} (\mu_{k,m} - \mu_{k})^2.
\end{equation*}

\subsection{Uncertainty-Aware Adaptation with Experience Recalling}\label{sc:mainlearn}
To adapt the traversability prediction model across diverse environments, we leverage a generative model–based continual learning framework. The generative model recalls past input–output samples for experience replay without storing raw data. Before delving into the details of our continual learning framework, the initial step is established as follows.

\subsubsection{Initial training of the traversability prediction model}
we begin with the initial model, $\mathcal{T}^{1}$, trained on the first environment dataset $\mathcal{D}^{1}$, which serves as the foundation for adaptation. This initial model $\mathcal{T}^{1}$ is trained by minimizing the Negative Log-Likelihood (NLL) loss, denoted as $\mathcal{L}_{\text{NLL}}$, which captures the likelihood of the estimated traction parameters given the corresponding terrain features and the robot's dynamics states, i.e., velocity.

\subsubsection{Initial training of the generative experience recall model} we design a generative recall model $\mathcal{R}^{1}$, which randomly synthesizes the prediction model's input and output data, denoted as $\mathcal{D}^{\text{recall}}$. As with the prediction model, the input data consists of two components: terrain features $\phi^{\text{recall}}$ and robot dynamic states, i.e., $v^{\text{recall}}$. Leveraging the fact that velocities are physical quantities with bounded ranges, we design a conditional recall process that first samples velocity and then generates corresponding terrain features conditioned on it. To accomplish this, we adopt a Conditional Variational Autoencoder (CVAE) \cite{sohn2015learning}, which consists of an encoder and a decoder, denoted as $q_{enc}(z|\phi,v)$ and $ p_{dec}(\phi|v,z)$, respectively. Upon training of the CVAE, the decoder is used to recall features $\phi^{\text{recall}}$ conditioned on a sampled velocity state $v^{\text{recall}}$ and latent variable $z$. During recall, both $v^{\text{recall}}$ and $z$ are randomly sampled, where $v^{\text{recall}}$ respects the velocity limits defined by the robot’s hardware specifications, and $z$ is drawn from a Gaussian prior distribution. Next, to synthesize the output data, the trained traversability prediction model $\mathcal{T}$ processes the recalled inputs as follows:
\begin{equation*}
    \xi^{\text{recall}} \sim \mathcal{N}(\mu^{\text{recall}}, (\sigma^{\text{recall}})^2),
\end{equation*}
where $[\mu^{\text{recall}}, (\sigma^{\text{recall}})^2] = \mathcal{T}^{1}(\phi^{\text{recall}}, v^{\text{recall}}).$ This prediction includes $\sigma^{\text{recall}}$ as a measure of uncertainty, reflecting the model’s confidence in its generated samples. Specifically, a higher $\sigma^{\text{recall}}$ indicates that the recalled sample likely originates from an environment that was not well represented in $\mathcal{D}^{1}$ for training $\mathcal{T}^1$.

\subsubsection{Uncertainty-aware traversability prediction model update}
we update the traversability prediction model from $\mathcal{T}^{i-1}$ to $\mathcal{T}^{i}$ for $i \geq 2$, using only the collected dataset $\mathcal{D}^i$ and the generative recall model $\mathcal{R}^{i-1}$. First, we generate a batch of recalled input-output samples using 
$\mathcal{R}^{i-1}$, overwriting the dataset $\mathcal{D}^{\text{recall}}$. These recalled samples are then combined with the newly collected dataset $\mathcal{D}^i$ to update the model $\mathcal{T}^{i}$ by minimizing the following loss function:
\begin{equation}\label{eq:trav_update}
\mathcal{L}_{\mathcal{T}^{i}} = \mathcal{L}_{\text{NLL}}(\mathcal{D}^i) + \lambda \mathcal{L}_{\text{adapt}}(\mathcal{D}^{\text{recall}}), 
\end{equation}
where $\lambda$ is a scaling constant. The first term, 
$\mathcal{L}_{\text{NLL}}(\mathcal{D}^i)$, captures the experience obtained from newly acquired data. The second term, $\mathcal{L}_{\text{adapt}}(\mathcal{D}^{\text{recall}})$, aims to retain prior experience through recalled samples that imitate past data. This is achieved by aligning the distributions of the predicted and recalled output distributions using the Jensen–Shannon (JS) divergence as a distance measure, defined as follows:
\begin{equation*} 
\begin{split}
    \mathcal{L}_{\text{adapt}}(\mathcal{D}^{\text{recall}})  :=  \sum_{\mathcal{D}^{\text{recall}}} &\frac{1}{2}KL\left(\mathcal{T}^{i}(\phi^{\text{recall}},v^{\text{recall}}) | \xi^{\text{recall}}\right) \\
    & + \frac{1}{2}KL\left(\xi^{\text{recall}} | \mathcal{T}^{i}(\phi^{\text{recall}},v^{\text{recall}}) \right),
\end{split}
\end{equation*}
where $KL$ denotes the Kullback-Leibler (KL) divergence. This distributional alignment alleviates catastrophic forgetting of recalled samples and mitigates overfitting to unreliable samples with higher $\sigma^{recall}$, thereby reducing the risk of erroneous updates.

\subsubsection{Uncertainty-aware generative experience recall model update}\label{sssc:uncrecall}
similar to the prediction model update, we update the generative experience recall model from $\mathcal{R}^{i-1}$ to $\mathcal{R}^{i}$ using the newly collected dataset $\mathcal{D}^{i}$ and recalled dataset $\mathcal{D}^{\text{recall}}$, without explicit access to past data, i.e., $\mathcal{D}^{l-1}, \forall l\leq i$. Specifically, we recall data by randomly sampling input features, denoted as $(\phi^{\text{recall}},v^{\text{recall}})$, using $\mathcal{R}^{i-1}$. The quality of the recalled samples is assessed using their uncertainty $(\sigma^{\text{recall}})^2$, predicted by the preceding prediction model $\mathcal{T}^{i-1}$. These uncertainty estimates act as proxies for distributional similarity to real data, enabling us to identify and filter recalled samples that deviate from the real data distribution. To this end, we apply a variance threshold $\tau$ to exclude uncertain recalls, yielding the refined recall dataset:
\begin{equation*}
    \bar{\mathcal{D}}^{\text{recall}} := \left\{ (\phi^{\text{recall}}, v^{\text{recall}}) \,\middle|\, \sigma^{\text{recall}} < \tau \right\}.
\end{equation*}
We then define the augmented dataset as the union of the newly acquired data \( \mathcal{D}^i \) and the filtered recalled samples:
\begin{equation}\label{eq:filt}
    \mathcal{D}^{\text{aug}} := \mathcal{D}^i \cup \bar{\mathcal{D}}^{\text{recall}}.
\end{equation}
The generative experience recall model \( \mathcal{R}^{i} \) is trained using $\mathcal{D}^{\text{aug}}$, ensuring that only recalled samples with sufficiently low uncertainty contribute to retaining prior experience while recently acquired data facilitates adaptation to the new environment.

A comprehensive overview of the proposed continual learning framework \footnote{Code available at \href{https://github.com/HMCL-UNIST/Continual-Traversability-Learning}{https://github.com/HMCL-UNIST/Continual-Traversability-Learning}.} with uncertainty-aware adaptation is presented in Algorithm 1.

\begin{algorithm}[th]
\caption{Continual Learning Framework with Uncertainty-aware Adaptation}
\label{alg:iterative_adaptation}
\begin{algorithmic}
\State \textbf{Initialize}: Set iteration index $i = 1$.
\\
\textbf{Input}: Initial dataset $\mathcal{D}^1$.
\end{algorithmic}
\begin{algorithmic}[1]    
    \State  Initial train $\mathcal{T}^1$ and $\mathcal{R}^1$ (Sections IV.E.1 and IV.E.2).
    \While{New dataset $\mathcal{D}^{i+1}$ is acquired}
        \State Generate recalled dataset $\mathcal{D}^{\text{recall}}$ using $\mathcal{R}^i$ and $\mathcal{T}^i$.       
        \State  Update $\mathcal{T}^i \rightarrow \mathcal{T}^{i+1}$ with \eqref{eq:trav_update} (Section IV.E.3).
        \State Update $\mathcal{R}^i \rightarrow \mathcal{R}^{i+1}$ using $\mathcal{D}^{\text{aug}}$ in \eqref{eq:filt} \ \ \ \ \par \par (Section IV.E.4).
        \State $i = i + 1$.        
    \EndWhile
\end{algorithmic}
\end{algorithm}

\subsection{Navigation with Traversability Prediction}\label{sc:mainmpc}
Based on the updated traversability prediction model, the optimal off-road navigation can be formulated as an MPC problem. The MPC minimizes a cost function associated with predicted future states as outlined in Section \ref{ssc:navigation}. Specifically, the objective cost in \eqref{eq:obj} is defined as:
\begin{equation*}
\mathcal{C}(x_t,\mu_t,\sigma^{2}_t) = w_1 \text{Goal}(p^x_t, p^y_t) + w_2 \sigma_{t}^2 +w_3 |1-\mu_t|,
\end{equation*}
where \( \text{Goal}(p^x_t, p^y_t) \) encourages progress toward the goal over a finite horizon, and \( \sigma_{t}^2 \) penalizes regions of high predictive uncertainty stemming from limited environment knowledge and/or irreducible measurement noise. The slip cost term $|1 - \mu_t|$ encourages the predicted traction to approach 1, representing ideal alignment between the robot’s actual motion and the commanded input under the nominal dynamics. Notably, $\mu_t$ is not constrained to lie within $[0, 1]$, allowing the model to capture a wider range of robot-terrain interactions. The weights \( w_1, w_2, w_3, \in \mathbb{R}^+ \) are tunable scaling factors.

To solve for \eqref{eq:obj}, which is subject to stochastic dynamics $f$, we adopt a mean-propagation strategy for cost approximation, following the method in \cite{lee2023kino}. Specifically, the robot dynamics are propagated using the mean estimates of uncertain traction parameters. The propagated states are then used to evaluate the cost function and optimize the control inputs. Under this strategy, the navigation problem can be expressed as the following MPC problem:
\begin{equation*}
\begin{split}
    \min\limits_{u_{0:T-1}} \ & \sum_{t=0}^{T}\mathcal{C}(x_t,\mu_t,\sigma^{2}_t)\\
     \mathrm{s.t.} & \ \  \quad x_{0}=x_{k}, \\
    &\ x_{t+1} = f(x_t, u_t, \mu_t),\  \forall t \in \{0,\cdots, T-1\}   \\ 
    & [\mu_t, \sigma_t^2] = \mathcal{T}(\phi_t, v_t), \forall t \in \{0,\cdots, T-1\}.
\end{split}
\end{equation*}

Upon optimization, only the first control $u_{0}$ is applied to the robot, while the remaining $u_{1:T-1}$ are temporarily stored in a buffer and used as fallback actions if the subsequent MPC computation fails to meet the desired control frequency.

\section{Field Experiment}

\subsection{Platform and Dataset Description}
The experiments are conducted on a Clearpath Jackal robot, equipped with an NVIDIA Jetson AGX Orin for onboard computation, an Ouster LiDAR, and a ZED X stereo camera to sense its surrounding environment. A Lidar-inertial odometry \cite{chen2023direct} provides high-frequency pose and velocity estimates. The ZED X camera captures both RGB images and depth point clouds, which are used to extract the terrain features. The extracted features are encoded into a multi-modal grid map with a resolution of $0.25$ meters \cite{erni23mem}, and represented as latent vectors of dimension $p=3$, updated at a rate of 20 Hz. An ensemble of $M=5$ models performs traversability prediction. The entire system is built within the ROS 2 environment. The control horizon is set to $T=30$ with a time step of $\Delta t = 0.1$ seconds, corresponding to a 3-second prediction window. A batch of 1024 rollouts is used per optimization cycle to support a control frequency of 10 Hz.

\begin{figure}[htpb]
\centering
\includegraphics[width=0.48\textwidth]{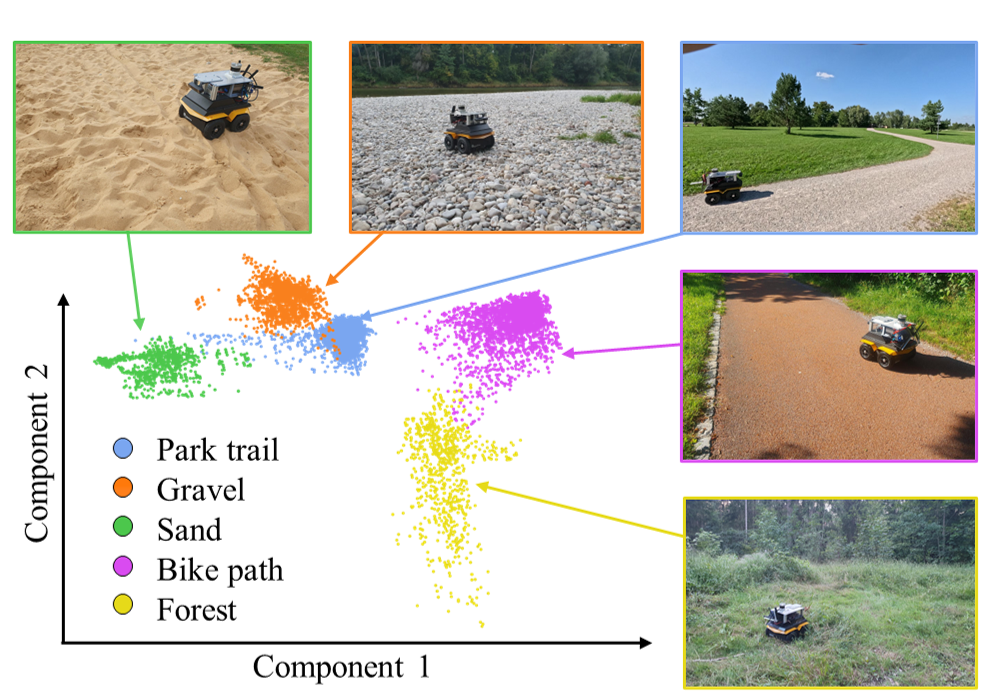}
\caption{PCA projection of terrain features $\phi$ from five distinct environments.}
\label{fig:pca_features}
\vspace{-3mm}
\end{figure}

The evaluation of continual learning necessitates diverse environments, which enables assessment of how well prior knowledge is retained when adapting to novel environments. In light of this requirement, we consider five environments with distinct terrain characteristics: an asphalt trail in a public park (Dataset $\mathcal{D}^1$), a gravel terrain near a river shore (Dataset $\mathcal{D}^2$), a sand area (Dataset $\mathcal{D}^3$), a bike path (Dataset $\mathcal{D}^4$), and a forest with dense vegetation (Dataset $\mathcal{D}^5$). At each site, we collected approximately 10 minutes of data by manually driving through both traversable and non-traversable areas. We apply Principal Component Analysis (PCA) to the extracted terrain features $\phi$, projecting them into a two-dimensional space to characterize distributional differences among the five environments, as illustrated in Figure~\ref{fig:pca_features}. The projected features reveal distinct domain shifts that are pertinent to a domain-incremental continual learning problem, as discussed in Section~\ref{ssc:problem_formulation}.

\subsection{Evaluation Metrics}
The traversability prediction model is trained sequentially on datasets $\mathcal{D}^1$ through $\mathcal{D}^5$, each corresponding to data collected from Environments 1 through 5. Continual learning performance is evaluated in terms of \textit{adaptation} and \textit{memory retention}. Both are computed from the prediction performance of the traction model on test datasets using the Negative Log-Likelihood (NLL). Let $n_{k,j} \in \mathbb{R},\  k \leq j$ denote the NLL on the test set of $\mathcal{D}^k$ after training on datasets $\mathcal{D}^{1:j}$. Adaptation performance at the environment $j$ is given by $n_{j,j}$, which measures how well the updated model $\mathcal{T}^j$ fits $\mathcal{D}^j$. A lower $n_{j,j}$ indicates better adaptation performance. Memory retention performance is evaluated using the Forgetting Measure (FM) \cite{wang2024comprehensive}, defined as $f_{k,j} = \max(0,\; n_{k,j} - \min_{i < j} n_{k,i})$, which captures the maximum performance degradation on $\mathcal{D}^k$ after training on $\mathcal{D}^j$. A lower FM indicates better experience retention and improved continual learning performance.

\subsection{Baseline Methods}
For comparative analysis, we evaluate the proposed continual learning framework against five baseline methods:
\begin{itemize}
    \item \textbf{Complete-Memory (CM):} This upper-bound baseline trains $\mathcal{T}^i$ using the full dataset $\mathcal{D}^j, \; 1\leq j \leq i$, enabling batch retraining with complete access to past data. 
    \item \textbf{Incremental-Memory (IMOST) \cite{ma2024imost}:}           
    This method updates $\mathcal{T}^{i-1}$ using an incremental dynamic memory buffer that stores new samples according to an information-expansion criterion.
    \item \textbf{Fast Adaptation (WVN) \cite{frey23fast}:}  
    This baseline updates $\mathcal{T}^{i-1}$ using only the recently acquired data $\mathcal{D}^i$, without considering experience retention.
    \item \textbf{LwF-fashion (LwF) \cite{li2017learning}:} This method uses the model $\mathcal{T}^{i-1}$ and the recently acquired data $\mathcal{D}^i$ to retrain $\mathcal{T}^i$ without accessing past data. Experience retention is achieved via distillation by minimizing the KL divergence between the outputs of $\mathcal{T}^{i-1}$ and $\mathcal{T}^i$ on $\mathcal{D}^i$.
    \item \textbf{Naive Generative Rehearsal (NGR):} A variant of our method that updates $\mathcal{T}^{i-1}$ with recalled samples $\mathcal{R}^{i-1}$ and the recently acquired data $\mathcal{D}^i$ without the uncertainty-aware adaptation in Section \ref{sssc:uncrecall}, serving as an ablation of the threshold $\tau$.
\end{itemize}
To ensure a fair comparison, all baseline methods are interfaced to take terrain and robot states as inputs and to output probabilistic traction, as in the proposed framework. For each method, we identify the hyperparameter setting on the Pareto front with equal weighting of adaptation and memory retention.

\subsection{Field Experiment Results and Discussion}
\subsubsection{Evaluation of experience recalling}
\begin{figure}[htpb]
\centering
\includegraphics[width=0.9\columnwidth]{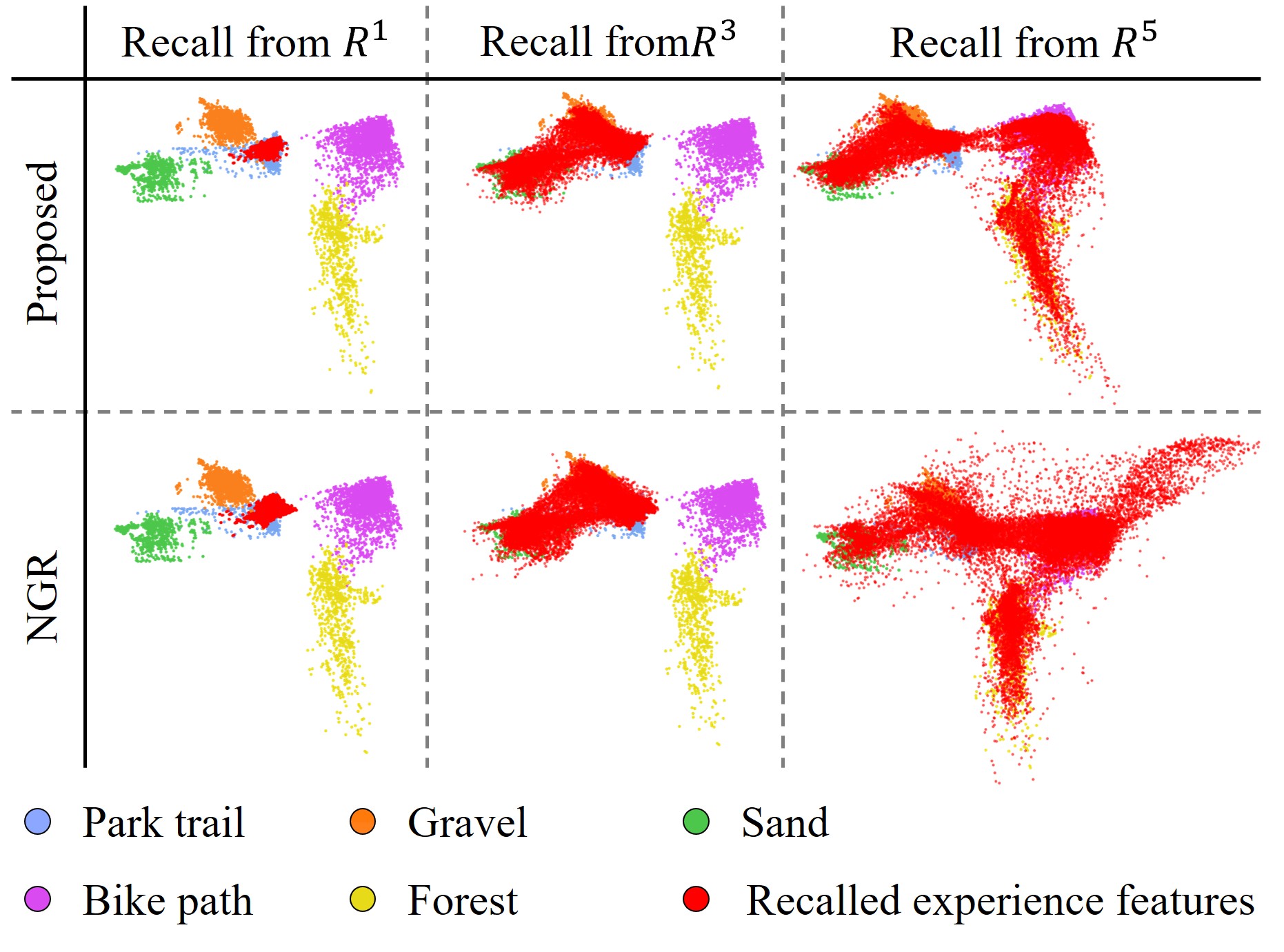}
\caption{PCA projection of recalled samples alongside real terrain features collected from Environments 1–5.}
\label{fig:recallpca}
\vspace{-3mm}
\end{figure}

we begin by assessing the validity of recalled samples generated by the generative experience recall model $\mathcal{R}^{i}$ during adaptation. The distributions of the recalled terrain features are visualized on top of the real terrain features using PCA, as shown in Figure~\ref{fig:recallpca}. While the proposed method produces samples that closely align with the real data distribution, the ablation method (NGR) generates dispersed samples that poorly align with real data. These results highlight that the uncertainty-aware adaptation introduced in Section~\ref{sssc:uncrecall} helps the recall model produce samples that better reflect collected experience, supporting more accurate model updates during continual learning.
\vspace{-3mm}

\subsubsection{Evaluation of adaptation performance}
as part of the quantitative evaluation, Table~\ref{tb:nll} and Table~\ref{tb:fm} report the mean and standard deviation of NLL and FM over ten training sessions for each continual learning scenario. The CM method consistently achieves the lowest forgetting, as it keeps access to the full dataset throughout training. Among those without direct access to past data, our proposed method yields the lowest FM, indicating improved retention of prior experience while effectively adapting to new environments. The IMOST method achieves retention performance comparable to, and occasionally exceeding, our method, but this comes at the cost of a continually growing memory requirement. This highlights the effectiveness of our uncertainty-aware adaptation for continual learning without storing past data, retaining prior experience with only the parameters of the prediction and recall models.

\begin{table}[htpb]
\centering
\caption{Adaptation statistics during continual learning (mean $\pm$ std).}
\setlength{\tabcolsep}{3pt}
\resizebox{\columnwidth}{!}{%
\scriptsize
\begin{tabular}{c|l|c|c|c|c}
\toprule
\multicolumn{2}{c|}{NLL, $n_{j,j}$} & \multicolumn{4}{c}{Training from $\mathcal{D}^1$ to $\mathcal{D}^j$} \\
\midrule
\multicolumn{2}{c|}{} &
$\mathcal{D}^1 \!\rightarrow\! \mathcal{D}^2$ &
$\mathcal{D}^1 \!\rightarrow\! \mathcal{D}^2 \!\rightarrow\! \mathcal{D}^3$ &
$\mathcal{D}^1 \!\rightarrow\! \cdots \!\rightarrow\! \mathcal{D}^4$ &
$\mathcal{D}^1 \!\rightarrow\! \cdots \!\rightarrow\! \mathcal{D}^5$ \\
\midrule
\multicolumn{2}{c|}{Test $j$} &
\textit{$\mathcal{D}^2$} &
\textit{$\mathcal{D}^3$} &
\textit{$\mathcal{D}^4$} &
\textit{$\mathcal{D}^5$} \\
\midrule
\multirow{6}{*}{Method} 
  & CM  & 0.17 $\pm$ 0.15 & \textbf{-0.61 $\pm$ 0.09} & \textbf{0.89 $\pm$ 0.13} & \textbf{-0.66 $\pm$ 0.06}\\
  & IMOST & 0.50 $\pm$ 0.26 & -0.46 $\pm$ 0.08 & 1.27 $\pm$ 0.07 & -0.35 $\pm$ 0.06\\
  & WVN        & \textbf{0.13 $\pm$ 0.08} & -0.57 $\pm$ 0.07 & 1.06 $\pm$ 0.01 & -0.44 $\pm$ 0.07  \\
  & LwF     & 0.19 $\pm$ 0.12 & -0.55 $\pm$ 0.07 & 1.03 $\pm$ 0.11 & -0.57 $\pm$ 0.06\\
  & NGR      & 0.82 $\pm$ 0.34 & -0.49 $\pm$ 0.06 & 1.20 $\pm$ 0.21 & -0.43 $\pm$ 0.02\\
  & Proposed & 0.71 $\pm$ 0.29 & -0.25 $\pm$ 0.11 & 0.97 $\pm$ 0.30 & -0.47 $\pm$ 0.05\\
\bottomrule
\end{tabular}
}\vspace{-3mm}
\label{tb:nll}
\end{table}

\begin{table}[htpb]
\centering
\caption{FM statistics during continual learning (mean $\pm$ std).}
\setlength{\tabcolsep}{2pt}
\resizebox{\columnwidth}{!}{%
{\scriptsize
\begin{tabular}{%
>{\centering\arraybackslash}m{0.65cm}|
>{\centering\arraybackslash}m{1.35cm}||
>{\centering\arraybackslash}m{1.35cm}|
>{\centering\arraybackslash}m{1.55cm}|
>{\centering\arraybackslash}m{1.55cm}|
>{\centering\arraybackslash}m{1.55cm}}
\toprule
\multicolumn{2}{c||}{FM, $f_{k,j}$} & \multicolumn{4}{c}{Training from $\mathcal{D}^1$ to $\mathcal{D}^j$ }\\
\midrule
Test $k$ & Method & $\mathcal{D}^1 \!\rightarrow\! \mathcal{D}^2$ &
$\mathcal{D}^1 \!\rightarrow\! \mathcal{D}^2 \!\rightarrow\! \mathcal{D}^3$ &
$\mathcal{D}^1 \!\rightarrow\! \cdots \!\rightarrow\! \mathcal{D}^4$ &
$\mathcal{D}^1 \!\rightarrow\! \cdots \!\rightarrow\! \mathcal{D}^5$ \\
\midrule
\multirow{6}{*}{$\mathcal{D}^1$}
& CM         & \textit{0.60 $\pm$ 0.29} & \textit{0.57 $\pm$ 0.12} & \textit{0.70 $\pm$ 0.20} & \textit{0.76 $\pm$ 0.17} \\
& IMOST     & 0.73 $\pm$ 0.10 & 1.12 $\pm$ 0.12 & 1.33 $\pm$ 0.11 & 1.55 $\pm$ 0.24 \\
& WVN      & 1.53 $\pm$ 0.01 & 2.08 $\pm$ 0.03 & 1.95 $\pm$ 0.11 & 2.56 $\pm$ 0.03 \\
& LwF & 1.60 $\pm$ 0.12 & 2.24 $\pm$ 0.23 & 2.06 $\pm$ 0.14 & 2.68 $\pm$ 0.16 \\
& NGR        & 0.80 $\pm$ 0.05 & 1.38 $\pm$ 0.08 & 1.42 $\pm$ 0.06 & 1.82 $\pm$ 0.21 \\
& Proposed   & \textbf{0.67 $\pm$ 0.05}   & \textbf{1.11 $\pm$ 0.07}   & \textbf{1.24 $\pm$ 0.08}   & \textbf{1.45 $\pm$ 0.13} \\ 
\midrule
\multirow{6}{*}{$\mathcal{D}^2$}
& CM         & --              & \textit{0.12 $\pm$ 0.12} & \textit{0.20 $\pm$ 0.14} & \textit{0.22 $\pm$ 0.09} \\
& IMOST     & --              & \textbf{0.38 $\pm$ 0.25} & \textbf{0.67 $\pm$ 0.19} & 1.03 $\pm$ 0.51 \\
& WVN      & --              & 2.17 $\pm$ 0.32 & 2.28 $\pm$ 0.39 & 2.29 $\pm$ 0.19 \\
& LwF& --              & 2.33 $\pm$ 0.34 & 2.26 $\pm$ 0.30 & 2.72 $\pm$ 0.18 \\
& NGR        & --              & 0.55 $\pm$ 0.15 & 0.68 $\pm$ 0.14 & 1.22 $\pm$ 0.37 \\
& Proposed   & --              & 0.38 $\pm$ 0.27   & 0.68 $\pm$ 0.18   & \textbf{0.98 $\pm$ 0.23} \\
\midrule
\multirow{6}{*}{$\mathcal{D}^3$}
& CM         & --              & --              & \textit{0.35 $\pm$ 0.17} & \textit{0.49 $\pm$ 0.20} \\
& IMOST     & --              & --              & 1.24 $\pm$ 0.23 & 1.51 $\pm$ 0.22 \\
& WVN      & --              & --              & 3.07 $\pm$ 0.33 & 3.18 $\pm$ 0.25 \\
& LwF & --              & --              & 2.85 $\pm$ 0.11 & 3.50 $\pm$ 0.45 \\
& NGR        & --              & --              & 1.39 $\pm$ 0.14 & 2.13 $\pm$ 0.19 \\
& Proposed   & --              & --              & \textbf{0.93 $\pm$ 0.21}   & \textbf{1.49 $\pm$ 0.31} \\
\midrule
\multirow{6}{*}{$\mathcal{D}^4$}
& CM         & --              & --              & --              & \textit{0.17 $\pm$ 0.12} \\
& IMOST     & --              & --              & --              & 0.36 $\pm$ 0.15 \\
& WVN      & --              & --              & --              & 1.60 $\pm$  0.21  \\
& LwF & --              & --              & --              & 1.52 $\pm$ 0.17 \\
& NGR        & --              & --              & --              & 0.29 $\pm$ 0.11 \\
& Proposed   & --              & --              & --              & \textbf{0.25 $\pm$ 0.16} \\
\bottomrule
\end{tabular}
}%
}\vspace{-3mm}
\label{tb:fm}
\end{table}

\begin{figure}[htpb]
\centering
\includegraphics[width=0.47\textwidth]{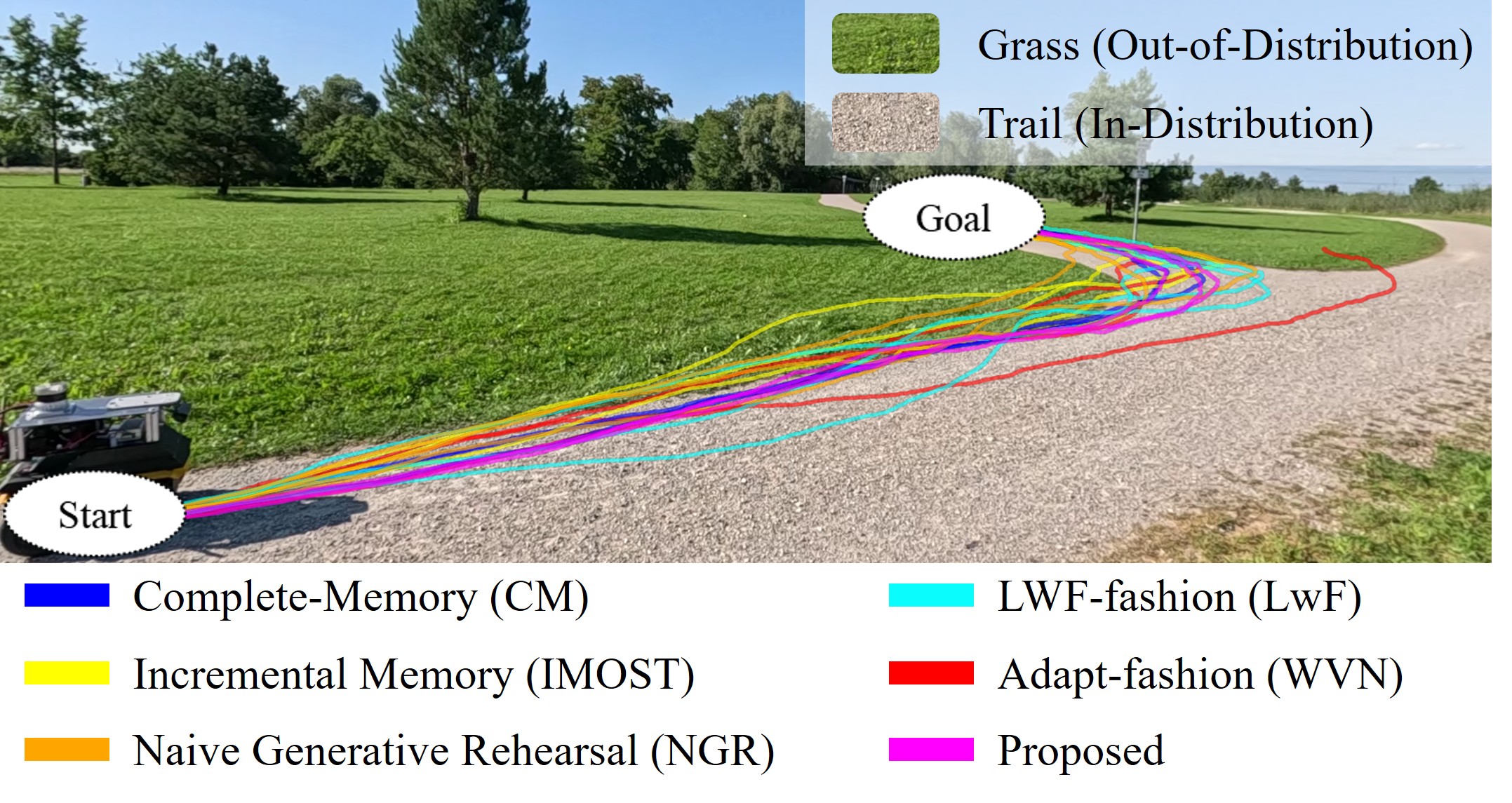}
\caption{Navigation results in Environment~1 using the traversability prediction model $\mathcal{T}^5$, with the paths of each method overlaid on the field images. }
\label{fig:nav_trials}
\end{figure}

Apart from the quantitative findings, we qualitatively evaluate the traversability prediction model’s ability to retain prior experience through navigation trials. Specifically, we assess the performance of model $\mathcal{T}^{5}$ in Environments 1 after sequential training on Environments 1 through 5 $($i.e., $ \mathcal{D}^1 \rightarrow \cdots \rightarrow \mathcal{D}^5)$. To probe the impact of uncertainty-aware navigation under continual learning, the grass region in the park (treated as an OoD area) was excluded from data collection, while only the asphalt region (in-distribution) was traversed during training. Figure~\ref{fig:nav_trials} illustrates representative trajectories from three trials per method, and Table~\ref{tb:navigation_results} summarizes the navigation results. Here, Goal reached denotes the number of successful goal completions out of three trials, and OoD avoidance indicates the number of trials in which the robot successfully avoided entering the grass region. The WVN method fails to reach the goal in one trial, reflecting degraded predictions due to forgetting. Other baselines, except CM and the proposed method, also fail to retain prior experience and incorrectly estimate uncertainty in the OoD region, resulting in poor avoidance performance. In contrast, the proposed method demonstrates stronger experience retention, reliably reaching the goal while avoiding OoD areas. Despite being trained without access to past data, it performs comparably to the CM method. These results demonstrate that our method effectively retains prior experience, enabling reliable navigation under continual learning scenarios.

\begin{table}[htpb]
\centering
\caption{Navigation results across 3 trials per method. Values indicate the number of successful trials.}
\setlength{\tabcolsep}{3pt}
\scriptsize
\begin{tabular}{l||c|c|c|c|c|c}
\toprule
\diagbox[width=12em]{Metric}{Method} & CM & IMOST & WVN & LwF & NGR & Proposed  \\
\midrule
Goal reached             & 3 & 3 & 2 & 3 & 3 & 3 \\
OoD avoidance  & 3 & 0 & 0 & 0 & 1 & 3 \\
\bottomrule
\end{tabular}
\label{tb:navigation_results}
\end{table}
\vspace{-2mm}

\section{Conclusions}
This paper presents a novel continual learning framework for traversability prediction in unstructured environments, addressing the challenges of adaptation and catastrophic forgetting without requiring explicit storage of past data. The core contribution is an uncertainty-aware adaptation strategy that selectively updates both the traversability prediction model and the generative experience recall model based on uncertainty estimates of recalled experiences. Real-world experiments with a skid-steering robot validate the effectiveness of the proposed method, demonstrating improved experience retention and reliable navigation performance. Future work will investigate more generalizable learning strategies that leverage prior knowledge, such as physics-informed foundation models, for traversability prediction.

\addtolength{\textheight}{-0cm}

\bibliographystyle{unsrt}
\bibliography{ref.bib}{}

\end{document}